\pdfoutput=1

\documentclass[11pt]{article}

\usepackage{ACL2023}

\usepackage{times}
\usepackage{latexsym}

\usepackage[T1]{fontenc}

\usepackage[utf8]{inputenc}

\usepackage{microtype}

\usepackage{inconsolata}
\usepackage{kotex}
\usepackage{multirow}
\usepackage{amsmath}
\usepackage{graphicx}
\usepackage{caption}
\usepackage{makecell}
\usepackage{hhline}
\usepackage{url}

\title{Painsight: An Extendable Opinion Mining Framework \\for Detecting Pain Points Based on Online Customer Reviews}

\author{Yukyung Lee$^1$, Jaehee Kim$^1$, Doyoon Kim$^1$, Yookyung Kho$^1$,Younsun Kim$^2$, Pilsung Kang$^1$ \\
    $^1$Korea University, Seoul, Republic of Korea \\ 
    $^2$LG Electronics Inc., Seoul, Republic of Korea \\
    \texttt{\small $^1$\{yukyung\_lee,jaehee\_kim, doyooni303, yookyung\_kho, pilsung\_kang\}@korea.ac.kr}\\ 
    \texttt{\small $^2$younsunny.kim@gmail.com}}
  
\begin{document}
\maketitle
\begin{abstract}
As the e-commerce market continues to expand and online transactions proliferate, customer reviews have emerged as a critical element in shaping the purchasing decisions of prospective buyers. Previous studies have endeavored to identify key aspects of customer reviews through the development of sentiment analysis models and topic models. However, extracting specific dissatisfaction factors remains a challenging task. In this study, we delineate the pain point detection problem and propose Painsight, an unsupervised framework for automatically extracting distinct dissatisfaction factors from customer reviews without relying on ground truth labels. Painsight employs pre-trained language models to construct sentiment analysis and topic models, leveraging attribution scores derived from model gradients to extract dissatisfaction factors. Upon application of the proposed methodology to customer review data spanning five product categories, we successfully identified and categorized dissatisfaction factors within each group, as well as isolated factors for each type. Notably, Painsight outperformed benchmark methods, achieving substantial performance enhancements and exceptional results in human evaluations.
\end{abstract}

\section{Introduction}
The thriving e-commerce market has rendered online customer reviews an indispensable factor in influencing the purchasing decisions of potential consumers \cite{zhu2010impact, kwahk2017effects, dellarocas2007exploring}. These reviews offer invaluable insights for businesses, empowering them to refine their products and services \cite{alibasic2021applying}. The analysis of copious customer reviews to comprehend customer needs and pain points is of paramount importance for augmenting service quality and heightening customer satisfaction \cite{plotkina2016delight, eslami2018effects, wu2020exploring, berger2020uniting}. Pain points pertain to specific difficulties or problems encountered by customers while utilizing a product or service \cite{lee2014pig}. These encompass emotional challenges arising from psychological demands and the incongruities between customers' actual perceptions and their expectations of products and services. Identifying and addressing pain points serves as a crucial initial step in enhancing the quality of products and services \cite{east2008measuring, ho2013effects, geetha2017relationship}. 

The increasing importance of customer reviews has spurred a wealth of research into comprehending their effects through the application of natural language processing methodologies, encompassing opinion mining, sentiment analysis, topic modeling, and keyword extraction \cite{eslami2018effects, heng2018exploring, de-geyndt-etal-2022-sentemo}. Nonetheless, the identification of specific attributes, such as pain points in customer review data, necessitates annotation tailored to each product and service. Owing to this constraint, prior studies have predominantly concentrated on devising methods for extracting keywords, a notion more expansive than pain points. \citet{wang2018impact} extracted attributes from product specifications, while \citet{klein-etal-2022-opinion} employed a BERT-based BIO tagging \cite{ramshaw-marcus-1995-text} model to extract aspect and opinion terms. More recently, researchers have shifted their focus toward aspect-based sentiment analysis \cite{bu-etal-2021-asap},  striving to analyze customers' opinions at the granular aspect level, as opposed to mere classification into being positive or negative \cite{zhang2022survey}. Moreover, \citet{wu2020exploring} utilized Latent Dirichlet Allocation (LDA) to differentiate review types and explore topic-related negative emotions.

Existing research has mainly focused on extracting keywords and aspect terms from reviews or simply analyzing review types; nonetheless, such methodologies exhibit limitations. Firstly, pain points are not only more specific but also contingent upon the product or service in question, in contrast to keywords or aspect terms. This is because customers may articulate disparate pain points utilizing identical keywords. Consequently, a comprehensive definition of pain points must be established prior to their detection \cite{forman2008examining,  de2014scenario, wang2016biclustering}. Secondly, even when equipped with a definition for pain points, the unique nature of pain points across products and services necessitates the ongoing annotation of new products, an endeavor that is both labor-intensive and costly \cite{saura2021user}. To overcome these limitations, an automated framework for the definition, extraction, and analysis of pain points from customer reviews is requisite, which is applicable to any product or service. In this study, we propose Painsight, an extendable opinion-mining framework for pain point detection, composed of a series of modules. Painsight implements a pipeline that conducts sentiment analysis and topic modeling using pre-trained language models, subsequently extracting pain points based on gradient-based attribution. When applied to customer reviews encompassing five categories of home appliances, Painsight effectively classified pain points emerging in diverse product groups. The extracted pain points exhibited substantial enhancements in performance, both quantitatively and qualitatively, in comparison to the results procured by the baseline model. The main contributions of this study can be summarized as follows:
\begin{itemize}
\setlength\itemsep{-0.2em}
\item We propose Painsight, an automated and scalable opinion-mining framework explicitly tailored for pain point analysis.
\item Painsight encompasses a comprehensive pipeline that executes sentiment analysis, topic modeling, and task-specific gradient-based attribution, drawing on a pre-trained language model.
\item Painsight demonstrates both quantitatively and qualitatively exceptional performance in the accurate identification of pain points concerning sentiment and topic across different product groups.
\end{itemize}

\begin{figure*}[t!]
\centering
\includegraphics[width=1\textwidth]{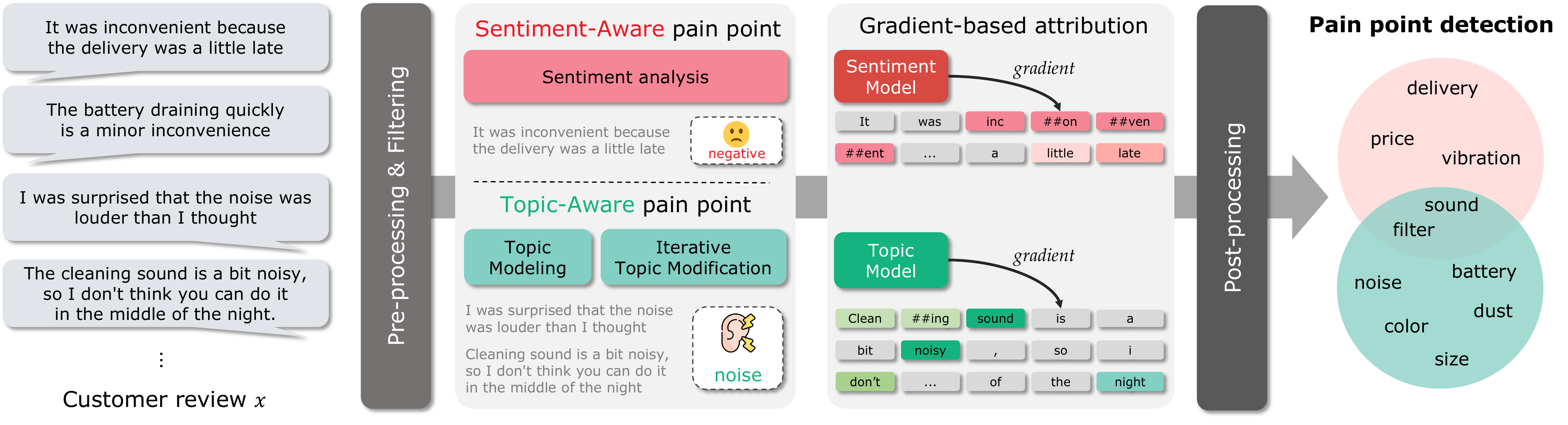}
\caption{Architecture of Painsight. Painsight receives customer reviews as input and constructs a sentiment-aware and topic-aware pain point model. The trained models are analyzed by a gradient-based attribution method to calculate the importance of each word in the input sentence, where darker tokens indicate a greater contribution to the prediction. The final pain points are detected through a series of post-processing steps based on high attribution scores.}
\label{fig:Painsight_Overview}
\end{figure*}

\section{Problem Statement} 
We aim to detect pain points in user-generated content (UGC), such as customer reviews, to identify customer discomforts and challenges \cite{cheng2021user}. However, the absence of standardized criteria for determining what constitutes pain points, contingent on the target product or service, poses a challenge because pain points exhibit variation in scale \cite{humphreys2018automated}. As a result, the precise definition of pain points pertinent to the target product or service is crucial.

\subsection{Definition of Pain Point}
\label{subsec:Definition of pain point}
Pain points arise from emotions customers experience while utilizing products and services, typically characterized by keywords reflecting negative opinions. However, not all negative keywords constitute pain points; rather, they signify complaints addressable through functional or procedural enhancements \cite{homburg2007see, rawson2013truth}. For instance, while "I tried vacuum A, and it's not good" conveys dissatisfaction without actionable insight, "I used vacuum B, and the battery drains too fast" distinctly identifies "battery" as an area for potential improvement. The scope of pain points can be determined based on their prevalence among users, with a focus on addressing common concerns to develop product and service improvement strategies, while simultaneously uncovering infrequent, personalized demands for diversification. In this study, we concentrate on detecting pain points affecting the majority of users, with the prospect of future methodology advancements catering to minority-specific pain points.

\subsection{Task Formulation}
\label{subsec:Task formulation}
We first formulate pain point detection, a novel task proposed in this study. Given $n$ customer review texts $\mathbf{x}=\left(\mathbf{x}_1, \mathbf{x}_2, \ldots, \mathbf{x}_n \right)$, the model aims to identify $k$ pain points ${p_1, \ldots, p_k}$ within a specific product group review dataset $\mathcal{X}$ ($\mathbf{x} \in \mathcal{X}$). The set of pain points in $\mathcal{X}$ is denoted as $\mathcal{P(X)}$, with each pain point $p_{k}$ $\in$ $\mathcal{P(X)}$ comprising contiguous tokens $ x_i, \ldots, x_{i+t}$ that form part of review text $\mathbf{x}$. In real-world settings, ground truth pain points for each product group are not predefined. Consequently, extracting suitable pain point candidates from review data and selecting product-specific pain points is essential. However, executing pain point extraction in a fully unsupervised manner presents significant challenges. Therefore, we assume the existence of relevant tasks with strong dependencies on pain point detection, with supervision $y$ provided for learning these tasks on the dataset. These relevant tasks serve as weak supervision, approximating pain point detection by modeling contexts of product-specific review and incorporating valuable knowledge. Given that each review is grounded in customer perception, words with substantial influence on relevant task predictions can be  understood as pain points of actual customers.

In this study, we utilized sentiment analysis and iterative topic modification (ITM) based on topic modeling as relevant tasks for pain point detection. Moreover, we aimed to extract keyword-based pain points using token attribution scores derived from these tasks. In both tasks, token attribution scores are computed through the following procedure. Given review $\mathbf{x}$ and its corresponding embedding sequence $\mathbf{e}=\left(\mathbf{e}_{1}, \mathbf{e}_{2}, \ldots, \mathbf{e}_{|\mathbf{x}|}\right)$, we define the relevant task classifier $f^{task}$, which takes the embedding sequence as input. The input gradient of each token can be used to evaluate the influence of the token for the target task, represented as a normalized gradient attribution vector $\mathbf{a}=\left(a_1, a_2, \ldots, a_{|\mathbf{x}|}\right)$. As described by \citet{wang-etal-2020-gradient}, the attribution at position $i$ can be expressed as:
\begin{align}
a_i=\frac{\left|\nabla_{\mathbf{x}_{i}} \mathcal{L} \cdot \mathbf{x}_{i}\right|}{\sum_j\left|\nabla_{\mathbf{x}_{j}} \mathcal{L} \cdot \mathbf{x}_{j}\right|}, \label{eq:attribution}
\end{align}
where $\mathcal{L}$ denotes the loss generated by classifier $f^{task}$, and $a_i$ is calculated through the dot product between the gradient of $\mathcal{L}$ and the embedding $\mathbf{e}_{i}$. Gradient-based attribution $\mathbf{a}$ \cite{sundararajan2017axiomatic, ijcai2017p371} represents each token's influence on the final prediction and can thus approximate word importance \cite{feng-etal-2018-pathologies}.

\section{Painsight}
The primary objective of the proposed framework is to model the entire process of automatically detecting pain points in real-world scenarios. Specifically, this study aims to address the following two practical research questions:
\begin{itemize}
\setlength\itemsep{-0.2em}
\item Q1: How do customers perceive products in general?
\item Q2: What types of discomfort do customers experience?
\end{itemize}
Considering the significance of both perspectives from prior works, a framework capable of generating accurate and diverse output, covering a wide range of distinct pain points, is necessary. The architecture of Painsight, depicted in Figure \ref{fig:Painsight_Overview}, features a parallel structure to incorporate these two research questions.

\begin{figure*}[t!]
\centering
\includegraphics[width=1\textwidth]{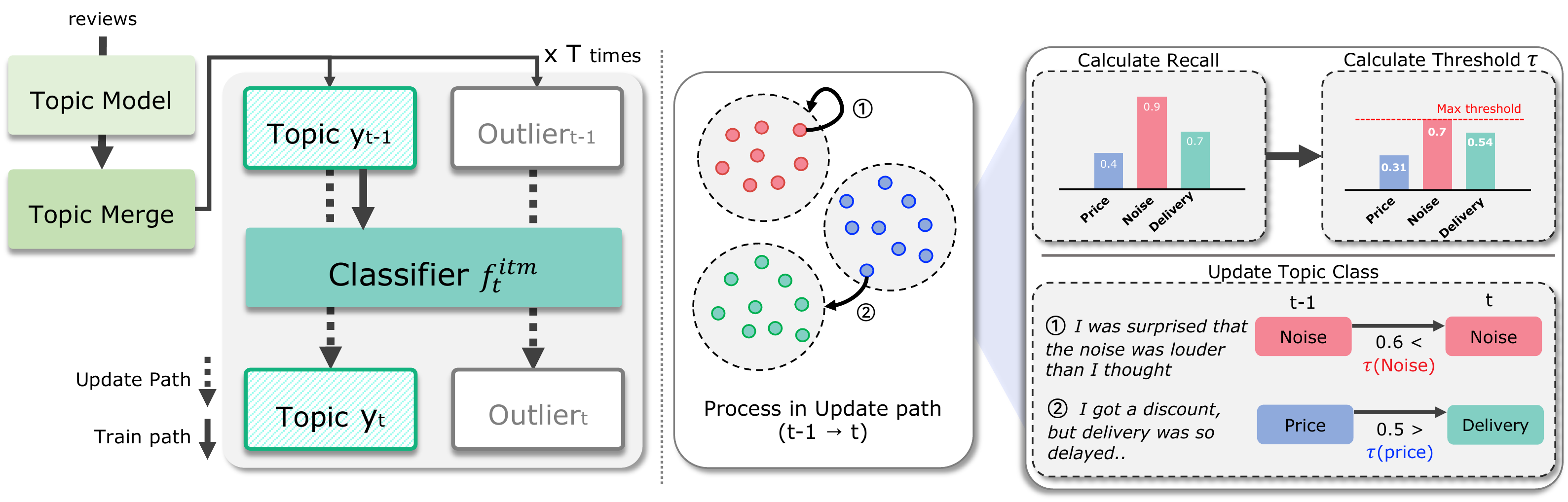}
\caption{Pipeline of topic-aware pain point. Review data undergoes topic modeling and merging to obtain initial topics. Then, a classifier $f_t^{itm}$ is trained for $T$ iterations. At each time step, recall is calculated based on predicted probabilities. Topic classes are updated when the confidence of class at time $t-$$1$ exceeds the threshold at time $t$.}
\label{fig:Topic-aware painpoint}
\end{figure*}

\subsection{Pre-processing \& Filtering}
We constructed a dataset by collecting various customer reviews from the web. As real-world customer reviews contain diverse noise, such as grammatical errors and outliers (i.e., irrelevant reviews), we applied three pre-processing steps for data refinement: \lowercase\expandafter{\romannumeral1}) spacing correction, \lowercase\expandafter{\romannumeral2}) keyword dictionary construction, and \lowercase\expandafter{\romannumeral3}) data filtering. Review data often consists of colloquial styles and may contain grammatical and spacing errors. We first utilized a pre-trained spacing model\footnote{We employed a character-level convolution neural network-based spacing model.} to correct spacing errors and employed regular expressions to fix common typos. Furthermore, as mentioned in Section \ref{subsec:Definition of pain point}, filtering out sentences without pain points is essential. To achieve this, we constructed a stopwords list and a keyword dictionary to filter out sentences expressing sentiments irrelevant to pain point extraction. We considered words frequently appearing in negative reviews and not included in the stopwords list as important keywords. Additionally, we filtered out duplicate reviews or those with fewer than ten tokens.

\subsection{Sentiment-Aware Pain Point}
\label{subsec:Sentiment-Aware pain point}
The Sentiment-Aware pain point approach focuses on negative emotions to address the question of \textbf{\textit{"How do customers perceive products in general?"}}. To achieve this goal, we conducted sentiment analysis on customer reviews using a classifier denoted as $f^{sent}$. The sentiment label $y$ is defined as $y \in \{positive, negative\}$, and $\hat{y}$ represents the predicted class, expressed as $\hat{y} = \operatorname{argmax}_{\hat{y}} f^{sent}(\hat{y} \mid \mathbf{x})$. Reviews predicted as `negative' by the trained $f^{sent}$ were considered likely to contain pain points.

\subsection{Topic-Aware Pain Point}
\label{subsec:Topic-Aware pain point}
The Topic-Aware pain point approach focuses on identifying specific types of discomfort that customers experience, addressing the question \textbf{\textit{"What types of pain points do customers encounter?"}}. We began by analyzing negative reviews to identify various types of complaints. To detect pain points by topic, we further refined sentences in outlier topics to properly segment them. The Topic-Aware pain point approach consists of three stages: \lowercase\expandafter{\romannumeral1}) topic modeling, \lowercase\expandafter{\romannumeral2}) topic merging, and \lowercase\expandafter{\romannumeral3}) ITM, as illustrated in Figure \ref{fig:Topic-aware painpoint}.

\textit{\textbf{Topic modeling}} The construction of Topic-Aware pain points necessitates establishing ground truth for each product group, which includes the number of topics and review-topic matching information. However, such labels are typically absent in customer reviews available on the web. Even when predefined pain points exist, detecting new pain points from real-time collected review data remains challenging. To address this issue, we first employed topic modeling to determine the initial topic class. Specifically, we utilized BERTopic \cite{grootendorst2022bertopic} to identify significant clustered topics. However, we observed that over $50\%$ of the data were classified as outliers, defined in this study as reviews devoid of any distinct topic, with more than 100 topics extracted. Misclassified topic modeling results can impede interpretability and provide users with incorrect pain points. To tackle this problem, we merged topics to relabel misclassified ones and performed ITM to accurately distinguish outlier reviews with low relevance to pain points.

\textit{\textbf{Topic merging}} In the above paragraph, we addressed the issue of reviews with similar topics being assigned to different clusters in the BERTopic output. To resolve this challenge, we propose a procedure for selecting representative words for each topic and determining which topics should be merged. Initially, we extracted nouns with the highest c-TF-IDF scores for each topic and designated them as representative words. Words with scores at least $s$ were considered keywords for the product group.\footnote{We set the hyperparameter $s$ to 0.1 based on experimental results.} Subsequently, we searched for topics containing these keywords and merged them into a single topic. Following the topic merging process, we observed that semantically similar topics were combined, resulting in roughly 20 to 30 merged topics. We also adjusted minor topic classes, which consisted of 5 to 10 reviews. However, the described topic merging approach primarily focuses on lexical matching assigned to the topics. Therefore, incorporating additional improvements that consider semantic aspects is crucial for refining the results.

\textit{\textbf{Iterative topic modification}} Despite the improvements in topic modeling results through merging, over $50\%$ of reviews still remain outliers. To address this problem, it is essential to assign initial topics that may be misclassified and to distinguish between reviews containing pain points and outlier reviews in the data. Consequently, we propose an ITM algorithm to enhance topic modeling results by updating the predicted topic with a confidence threshold above a certain level if it differs from the existing topic class during the training process. Our ITM algorithm is inspired by curriculum pseudo labeling (CPL) \cite{zhang2021flexmatch}, a method designed for semi-supervised learning that flexibly adjusts the threshold based on the learning difficulty for each class. Similar to CPL, the ITM algorithm adjusts the number of updated data points by varying the threshold according to the classification difficulty for each topic during the training process. As a result, this approach enables the modification of misclassified topic classes and induces additional topic merging.

Let the classifier for ITM be denoted as $f_{t}^{itm}$, where $t$ represents the current time-step of the classifier, and $f_{t}^{itm}\left(y_{t} \mid \mathbf{x}\right)$ represents the prediction probability of $y_{t}$. $y_t$ is the topic class at time $t$ for review $\mathbf{x}$ and belongs to the topic set $\mathcal{C}={topic_{1}, topic_{2},\ldots, topic_{M}}$. We used the merged topic output assigned to each review as the initial topic $y_{0}$. $\tau$ denotes a pre-defined threshold\footnote{$\tau$ is a hyperparameter set experimentally within the search space [0.4, 0.5, 0.6, 0.7].}, and the threshold, $\mathcal{T}(y_{t})$, for label modification based on $f_{t}^{itm}$ can be defined as:
\begin{align}
\mathcal{T}(y_{t})=\mathcal{R}(y_{t}) \cdot \tau. \label{eq:tau}
\end{align}
$\mathcal{T}({y}_{t})$ is a flexible threshold for topic class $y{t}$ at time-step $t$, and $\mathcal{R}(y_{t})$ is a relative recall value representing topic class difficulty. In CPL, accuracy is used as a measure of difficulty; however, accuracy can lead to biased estimates due to class imbalance in our topic modeling. As an alternative, we utilize recall, a sensitivity measure, to define the difficulty of the topic class. High recall indicates an easy class with high $\mathcal{T}(y)$, while low recall implies a difficult class with low $\mathcal{T}(y)$. Class difficulty is defined as:
\begin{align}
\mathcal{R}(y_{t})=
\frac{\operatorname{recall}(y_{t})}{\operatorname{max}_{y_{t} \in \mathcal{C}}(\operatorname{recall}(y_{t}))}. \label{eq:recall}
\end{align}
We train $f_{t}^{itm}$ to maximize log-likelihood based on the topic at $t-1$ and calculate the difficulty of each class $\mathcal{T}(y_{t})$ at every time-step\footnote{In this study, we set each epoch as a time-step.}. If the predicted probability $f_{t}^{itm}\left(y_{t} \mid \mathbf{x}\right)$ is greater than $\mathcal{T}(y_{t})$, we modify the topic to $y_{t}$. $y_{t}$ is defined as:
\begin{align}
    y_{t} = \Bigg\{
    \begin{array}{ll}
        y_{t} & \text{if } f_{t}^{itm}\left(y_{t} \mid \mathbf{x}\right)>\mathcal{T}(y_{t}),\\
        y_{t-1}, & \text{otherwise.}
    \end{array} \label{eq:y label}
\end{align}
Upon initializing the training with initial topics, we terminated the process when topics no longer merged or reached a satisfactory state\footnote{Meaning two out of three evaluation metrics no longer show improvement.}. We then considered the final prediction of ITM as the topic for each review.

\subsection{Gradient-based Attribution}
In this study, we aim to extract word importance related to pain points from two interdependent tasks: sentiment analysis and topic modeling. We employed gradient-based attribution, using token attribution scores for each task's prediction. Normalized attribution vectors $a_i^{sent}$ and $a_i^{itm}$ for individual tokens are derived from trained classifiers $f^{sent}$ and $f^{itm}$, as shown in Eq. (\ref{eq:attribution}):
\begin{align}
a_i^{sent}=\frac{\left|\nabla_{\mathbf{x}_{i}} \mathcal{L}_{sent} \cdot \mathbf{x}_{i}\right|}{\sum_j\left|\nabla_{\mathbf{x}_{j}} \mathcal{L}_{sent} \cdot \mathbf{x}_{j}\right|}, \label{eq:sent_attribution} \\
a_i^{itm}=\frac{\left|\nabla_{\mathbf{x}_{i}} \mathcal{L}_{topic} \cdot \mathbf{x}_{i}\right|}{\sum_j\left|\nabla_{\mathbf{x}_{j}} \mathcal{L}_{topic} \cdot \mathbf{x}_{j}\right|}. \label{eq:topic_attribution}
\end{align}
We computed $a_i^{sent}$ for `negative' reviews and $a_i^{itm}$ for reviews in $M-1$ topics, excluding outlier topics. We then selected the top-$g$ $a_i^{sent}$ and $a_i^{itm}$ with the highest attribution vectors and extracted words with index $i$. These words form the pain point candidate set $\mathcal{C(X)}$. The pain point set $\mathcal{P(X)}$ is a subset of $\mathcal{C(X)}$. We employed Captum \cite{kokhlikyan2020captum}'s integrated gradient method for axiomatic attribution calculation.

\subsection{Post-Processing}

The derived $\mathcal{C(X)}$, when used as the final result for pain points, has certain limitations. Since $a_i^{sent}$ and $a_i^{itm}$ are token attribution scores contributing to each class prediction, they tend to focus on words describing "states," such as verbs and adjectives like "not good" and "bad." To include both the subject and object related to the predicate's action, additional post-processing is necessary. We addressed this issue by refining the results using a dependency parsing model \cite{pororo}, a method that captures relationships between words in a sentence.

\begin{figure}[t!]
\centering
\includegraphics[width=1\columnwidth]{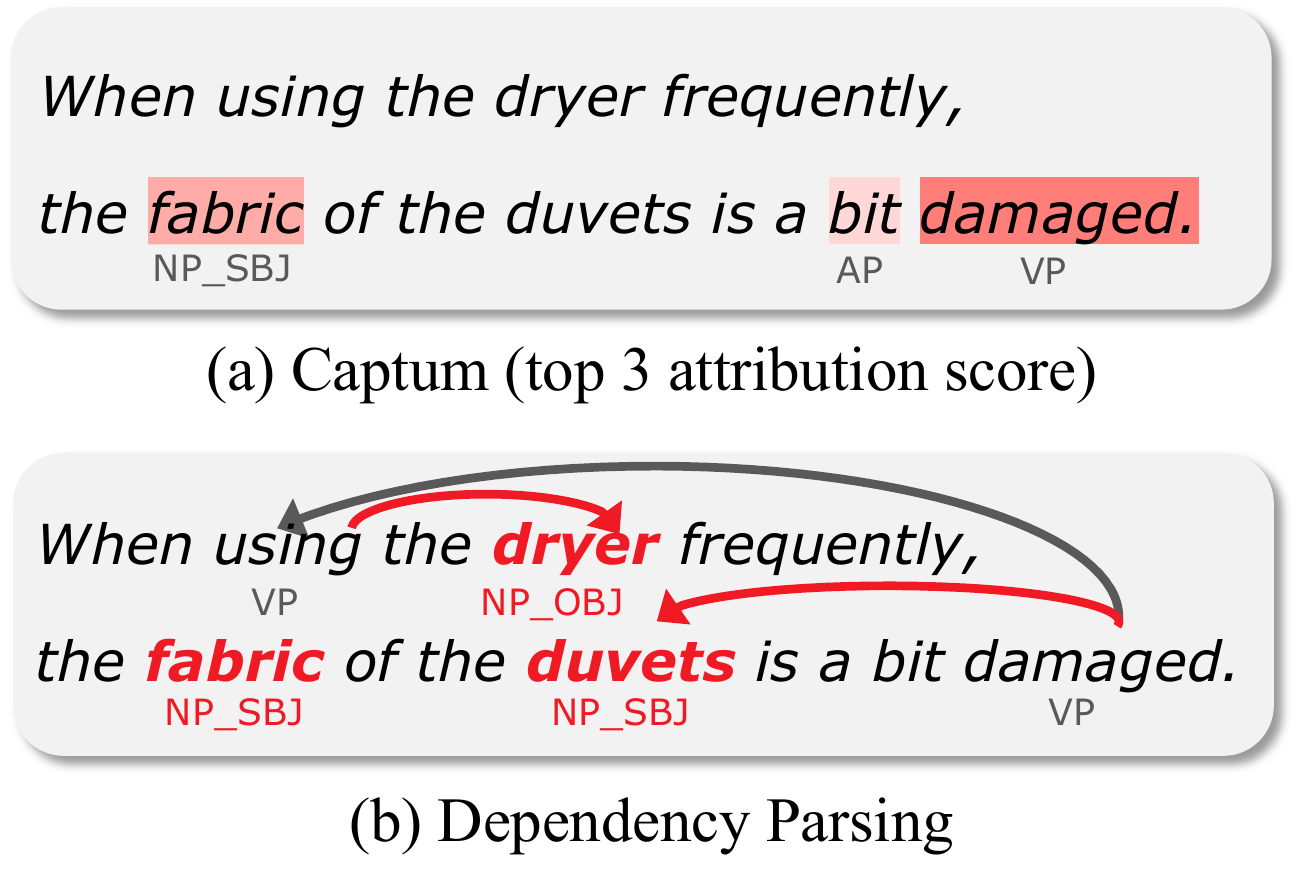}
\caption{Example of post-process. (a) represents the top three words with high attribution scores, where darker colors indicate a greater contribution to the prediction. (b) shows the result of dependency parsing, based on the part-of-speech information obtained in (a), to extract pain point candidates. The words highlighted in red belong to $\mathcal{C(X)}$.}
\label{fig:post process}
\end{figure}

First, we sorted the top $g$ tokens with the highest attribution scores from $f_{task}$ and extracted the corresponding words $w_i$. Then, we analyzed word relationships within sentences using dependency parsing. If $w_i$ belongs to a noun phrase (NP), we define it as $\mathcal{C(X)^{\prime}}$; if it belongs to a verb phrase (VP), we add the related NP to $\mathcal{C(X)^{\prime}}$. If the related word is a VP, we search for an NP to supplement $\mathcal{C(X)^{\prime}}$. We calculate word frequencies in the stopword-filtered $\mathcal{C(X)^{\prime}}$ and define the top $N$ words as the pain points $\mathcal{P(X)}$ for each product group\footnote{We set $g$ to 3, the number of $\mathcal{P(X)}_[{sent}$ to 30, and the number of $\mathcal{P(X)}_{topic}$ to 10 for each topic.}. Examples of post-processing can be found in Figure \ref{fig:post process}.

\section{Experiment setup}

\subsection{Dataset}
\begin{figure}[t!]
\centering
\includegraphics[width=1\columnwidth]{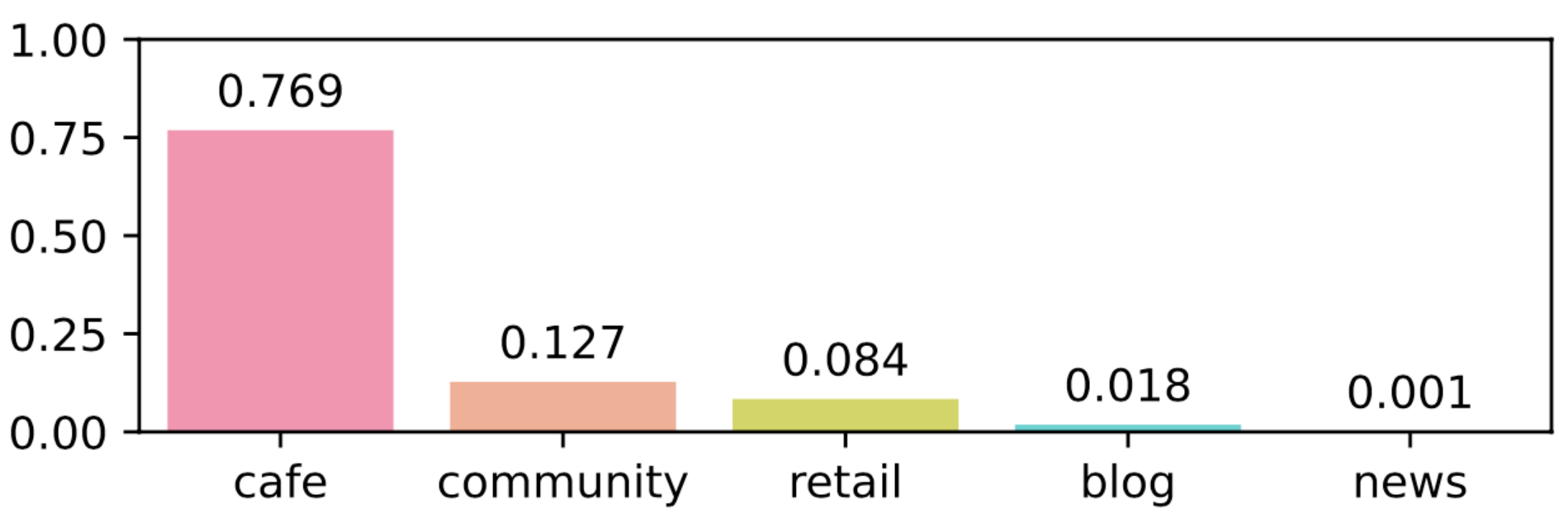}
\caption{The overall data distribution by site type.}
\label{fig:site type}
\end{figure}

We utilized Korean customer reviews on five home appliance categories gathered from various web sources between January 2020 and November 2021. The target categories included dryers, stylers, washing machines, vacuum cleaners, and robotic vacuums, obtained from 17 sites spanning five source types: retail, blog, cafe, community, and news. The dataset, originally collected by a home appliance company in Korea, had sentiment labels assigned based on the company's internal sentiment analysis logic within their voice of customer (VOC) analysis system. Data distribution and statistics for each product group are illustrated in Figure \ref{fig:site type} and Table \ref{tab:data statistic}. Detailed distribution of product categories can be found in Appendix \ref{app:data distribution}.

\begin{table}[t!]
\centering
\resizebox{\columnwidth}{!}{%
\setlength{\tabcolsep}{3pt}
\renewcommand{\arraystretch}{1.4}
\begin{tabular}{lccccc} 
\Xhline{3\arrayrulewidth}
                  & \multicolumn{1}{c}{\textbf{Dryer}} & \multicolumn{1}{c}{\textbf{Styler}} & \multicolumn{1}{c}{\textbf{WM}} & \multicolumn{1}{c}{\textbf{VC}} & \multicolumn{1}{c}{\textbf{RV}}  \\ 
\hline
\textbf{Positive} & 48,249                              & 26,453                               & 111,603                                       & 159,222                                      & 36,091                                        \\
\textbf{Negative} & 16,608                              & 9,034                                & 23,510                                        & 49,919                                       & 9,566                                         \\ 
\hline
\textbf{Total}    & 64,857                              & 35,487                               & 135,113                                       & 209,141                                      & 45,657                                        \\
\Xhline{3\arrayrulewidth}
\end{tabular}}
\caption{Data statistics for product groups.}
\label{tab:data statistic}
\end{table}

\subsection{Baselines}

\begin{table*}[t!]
\centering
\resizebox{\textwidth}{!}{%
\setlength{\tabcolsep}{1pt}
\renewcommand{\arraystretch}{1.3}
\setlength{\tabcolsep}{6pt}
\begin{tabular}{l|cc|cc|cc|cc|cc||cc} 
\Xhline{3\arrayrulewidth}
\multirow{2}{*}{\textbf{Model}} & \multicolumn{2}{c|}{\textbf{Dryer}} & \multicolumn{2}{c|}{\textbf{Styler}} & \multicolumn{2}{c|}{\textbf{WM}} & \multicolumn{2}{c|}{\textbf{VC}} & \multicolumn{2}{c||}{\textbf{RV}} & \multicolumn{2}{c}{\textbf{Avg. }}  \\ 
\cline{2-13}
                                & \textbf{Acc}   & \textbf{F1}        & \textbf{Acc}   & \textbf{F1}         & \textbf{Acc}   & \textbf{F1}                  & \textbf{Acc}   & \textbf{F1}                 & \textbf{Acc}   & \textbf{F1}                  & \textbf{Acc}   & \textbf{F1}       \\ 
\hline
\textbf{TextCNN}                & 94.35          & 92.76              & 90.20          & 84.32               & 96.75          & 94.31                        & 96.47          & 95.15                       & 95.28          & 93.06                        & 94.61          & 91.92             \\
\textbf{HAN}                    & 94.52          & 92.77              & 92.39          & 89.86               & 96.59          & 94.13                        & 96.14          & 94.74                       & 95.54          & 93.42                        & 95.04          & 92.98             \\
\textbf{BERT}                   & 98.40          & 97.93              & \textbf{97.57} & \textbf{96.80}      & 98.80          & 97.92                        & \textbf{98.36} & \textbf{97.70}              & 98.03          & 97.02                        & 98.23          & 97.47             \\
\textbf{BERT+AVocaDo}           & 97.52          & 96.79              & 96.23          & 95.01               & 98.55          & 97.50                        & 98.13          & 97.46                       & 98.39          & 97.58                        & 97.76          & 96.87             \\
\textbf{BERT+Vocab Expanision}  & \textbf{98.44} & \textbf{97.96}     & 97.23          & 96.35               & \textbf{98.81} & \textbf{97.94}               & 98.32          & \textbf{97.70}              & \textbf{98.52} & \textbf{97.63}               & \textbf{98.26} & \textbf{97.52}    \\
\Xhline{3\arrayrulewidth}
\end{tabular}}
\caption{Results on sentiment analysis in accuracy and f1 score. we utilize the following abbreviations: "WM" for Washing Machine, "VC" for Vacuum Cleaner, and "RV" for Robotic Vacuum. The average performance for the five datasets is presented in the rightmost column, with the highest performance indicated in bold.}
\label{tab:Sentiment Analysis Result}
\end{table*}

\subsubsection{Sentiment Analysis}

\textbf{TextCNN} \cite{kim-2014-convolutional} utilizes 1D convolutions to capture variable-length local features and n-grams.

\noindent \textbf{Hierarchical Attention Network (HAN)} \cite{yang2016hierarchical} consists of a single-layer Gated Recurrent Unit (GRU) that utilizes attention mechanisms at the word, sentence, and document levels to extract important information effectively.

\noindent\textbf{BERT} \cite{devlin-etal-2019-bert}is a bidirectional pre-trained language model \footnote{The `klue/bert-base' version was employed.}.

\noindent\textbf{BERT + AVocaDo} \cite{hong-etal-2021-avocado} is an extension of BERT with AVocaDo technique. It is designed to handle product-specific review data by performing vocab expansion on a domain-specific corpus. The expansion process is based on tokenizer training and calculation of the sub-word segmentation ratio for the domain corpus, with contrastive learning between the original tokenizer and the expanded one during fine-tuning.

\noindent\textbf{BERT + Vocab Expansion} manually adds frequently occurring vocabulary from the review to the BERT model. High-frequency words were identified using a count-based approach, and those not present in BERT vocabulary were added.

\subsubsection{Topic Modeling}

\textbf{LDA} \cite{blei2003latent} is a method inferring topics from word occurrence patterns in a corpus. While LDA provides multiple topics for each review, for comparison with Painsight, we considered the highest probability topic as the representative.

\noindent\textbf{BERTopic} is a BERT-based topic model \footnote{In this study, we used the "sentence-transformers/xlm-r-100langs-bert-base-nli-stsb-mean-tokens" model.} to generate document embeddings, clustering and creates representations using the c-TF-IDF procedure.

\subsubsection{Pain Point Detection}

\textbf{spaCy} \cite{spaCy2} is baseline using part-of-speech tagging to extract nouns within sentences. The same post-processing method as Painsight was applied.

\begin{table}[t!]
\centering
\resizebox{0.9\columnwidth}{!}{%
\setlength{\tabcolsep}{10pt}
\renewcommand{\arraystretch}{1.3}
\begin{tabular}{c|l|cccc} 
\Xhline{3\arrayrulewidth}
\multicolumn{1}{l|}{\textbf{Dataset}} & \textbf{Model} & \textbf{NPMI} $\uparrow$  & \textbf{Outlier (\%)}  \\ 
\hline
\multirow{3}{*}{\textbf{Dryer}}       & LDA            & -0.0327         & -                      \\
                                      & BERTopic       & \textbf{0.0558} & 59.39\%                \\
                                      & ITM (Ours)     & 0.0208          & \textbf{30.10}\%       \\ 
\hline
\multirow{3}{*}{\textbf{Styler}}      & LDA            & -0.0470         & -                      \\
                                      & BERTopic       & 0.0379          & 55.81\%                \\
                                      & ITM (Ours)     & \textbf{0.0737} & \textbf{39.39\%}       \\ 
\hline
\multirow{3}{*}{\textbf{WM}}          & LDA            & -0.0055         & -                      \\
                                      & BERTopic       & 0.0406          & 50.53\%                \\
                                      & ITM (Ours)     & \textbf{0.0487} & \textbf{29.92\%}       \\ 
\hline
\multirow{3}{*}{\textbf{VC}}          & LDA            & 0.0244          & -                      \\
                                      & BERTopic       & 0.0512          & 71.92\%                \\
                                      & ITM (Ours)     & \textbf{0.0628} & \textbf{26.67\%}       \\ 
\hline
\multirow{3}{*}{\textbf{RV}}          & LDA            & -0.0232         & -                      \\
                                      & BERTopic       & \textbf{0.0488} & 56.39\%                \\
                                      & ITM (Ours)     & 0.0450          & \textbf{37.81\%}       \\
\Xhline{3\arrayrulewidth}
\end{tabular}}
\caption{Results on topic modeling in NPMI, The rightmost column shows the proportion of outlier topics for each method, except for LDA, which does not extract outlier topics separately. The highest performance and low outlier ratio are marked in bold.}
\label{tab:Iterative Topic Modification}
\end{table}

\section{Experiment Results}
\subsection{Sentiment Analysis}
Table \ref{tab:Sentiment Analysis Result} presents the sentiment analysis performance across the five product groups. Given that review data is domain-specific, the language model can benefit from a vocabulary expansion process \cite{hong-etal-2021-avocado}. We aimed to enhance performance by applying AVocaDo and Vocab Expansion to the original BERT model. The experimental results showed that the BERT + Vocab Expansion model achieved superior performance and was thus selected as the final model for calculating attribution scores.

\subsection{Topic Modeling}
Table \ref{tab:Iterative Topic Modification} presents the topic modeling performance for the five product groups. We evaluated the performance of LDA, BERTopic, and ITM using coherence-based clustering metrics such as NPMI. LDA generates the desired number of topics without considering outliers, resulting in significantly lower performance compared to BERTopic and ITM. Conversely, BERTopic designates, on average, 57\% of the reviews as outliers and assigns topics only to the remaining data. This results in BERTopic being evaluated with a smaller number of reviews compared to other methods, which could create a more favorable environment for BERTopic's performance measurements. In contrast, our ITM algorithm effectively assigns reviews initially deemed as outliers to suitable topics, achieving comparable or superior performance to BERTopic in most cases. These results demonstrate that ITM successfully reduced the average proportion of outliers to 32\%, even under relatively challenging experimental conditions.

\subsection{Pain Point Detection}

As discussed in Section \ref{subsec:Task formulation}, real-world customer reviews lack gold labels for pain points. Thus, to validate the effectiveness of Painsight in detecting sentiment-aware and topic-aware pain points, we conducted human evaluations. We designed experiments for three tasks, each addressing the following research questions:
\begin{itemize}
\setlength\itemsep{-0.2em}
\item Task 1: Can the sentiment-aware pain point module extract appropriate pain point candidates from each sentence?
\item Task 2: Can the sentiment-aware pain point module identify suitable pain points for each product category?
\item Task 3: Can the topic-aware pain point module detect appropriate pain points for specific topics within each product category?
\end{itemize}

We conducted human evaluations for the three tasks across five product groups, with eight unique human judges participating in each experiment. In Task 1, we randomly selected 100 sentences per product category and used spaCy and Painsight to identify pain point candidates in each sentence. Judges assessed the appropriateness of the extracted candidates on a scale from 1 to 5, where 1 signifies \texttt{`all words are extracted incorrectly'} and 5 indicates \texttt{`all words are extracted correctly'}. In Task 2, we provided 100 randomly sampled example sentences for each product category and examined the adequacy of the final pain points identified for each category. Judges evaluated each pain point word, assigning a score of 0 for unsuitable and 1 for suitable. For Task 3, we randomly selected 20 example sentences per product category according to topics and assessed the appropriateness of the detected pain points for the corresponding topics. Judges assigned a score of 0 for unsuitable and 1 for suitable pain point words.

\begin{table}[t!]
\centering
\resizebox{\columnwidth}{!}{%
\setlength{\tabcolsep}{3pt}
\renewcommand{\arraystretch}{1.3}
\setlength{\tabcolsep}{6pt}
\begin{tabular}{lccccc|c} 
\Xhline{3\arrayrulewidth}
\multirow{2}{*}{\textbf{4 - (a)}} & \multicolumn{6}{c}{\textit{\textbf{Task 1 (1\textasciitilde{}5 scale)}}}                                   \\ 
\cline{2-7}
                       & \textbf{Dryer} & \textbf{Styler} & \textbf{WM}   & \textbf{VC}   & \textbf{RV}   & \textbf{Avg.}  \\ 
\hline
spaCy                  & 3.02           & 2.93            & 2.67          & 2.99          & 3.17          & 2.95           \\
Painsight              & \textbf{3.64}  & \textbf{3.57}   & \textbf{3.41} & \textbf{3.58} & \textbf{3.81} & \textbf{3.60}  \\ 
\Xhline{2\arrayrulewidth}
\multirow{2}{*}{\textbf{4 - (b)}} & \multicolumn{6}{c}{\textit{\textbf{Task 2 (0 or 1)}}}                                                      \\ 
\cline{2-7}
                       & \textbf{Dryer} & \textbf{Styler} & \textbf{WM}   & \textbf{VC}   & \textbf{RV}   & \textbf{Avg.}  \\ 
\hline
spaCy                  & 0.53           & 0.61            & 0.45          & 0.54          & 0.60          & 0.55           \\
Painsight              & \textbf{0.72}  & \textbf{0.77}   & \textbf{0.67} & \textbf{0.69} & \textbf{0.75} & \textbf{0.72}  \\ 
\Xhline{2\arrayrulewidth}
\multirow{2}{*}{\textbf{4 - (c)}} & \multicolumn{6}{c}{\textit{\textbf{Task 3 (0 or 1)}}}                                                      \\ 
\cline{2-7}
                       & \textbf{Dryer} & \textbf{Styler} & \textbf{WM}   & \textbf{VC}   & \textbf{RV}   & \textbf{Avg.}  \\ 
\hline
LDA                    & 0.13           & 0.16            & 0.18          & 0.14          & 0.10          & 0.12           \\
BERTopic               & 0.29           & 0.31            & 0.40          & 0.33          & 0.32          & 0.33           \\
Painsight              & \textbf{0.54}  & \textbf{0.51}   & \textbf{0.49} & \textbf{0.39} & \textbf{0.51} & \textbf{0.47}  \\
\Xhline{3\arrayrulewidth}
\end{tabular}}
\caption{Results of human evaluation for Tasks 1, 2, and 3. Task 1 was evaluated on a scale of 1 to 5, while Tasks 2 and 3 were assessed with scores of 0 or 1. The average performance across the five datasets is displayed in the rightmost column, with the highest performance indicated in bold. Paired t-tests were conducted comparing baselines with Painsight, and all experiments exhibited significant differences with p-values < 0.001.} 
\label{tab:human evaluation}
\end{table}

Table \ref{tab:human evaluation} presents the human evaluation results across the three tasks. First, in Task 1 (Table \ref{tab:human evaluation} - (a)), which concentrates on extracting pain point candidates from each review sentence, Painsight's approach — extracting words contributing to `negative' predictions based on high attribution scores — outperforms spaCy's noun extraction, with an average improvement of 0.65 points. In Task 2 (Table \ref{tab:human evaluation} - (b)), the performance of the final sentiment-aware pain points is evaluated. These pain points are obtained by sorting the pain point candidates from Task 1 by frequency across all product categories. Assessing the output for each word reveals an average improvement of 0.17 points across the five product categories. Lastly, Task 3 (Table \ref{tab:human evaluation} - (c)) examines the final results of the topic-aware pain points by extracting pain points for each topic within the product categories. This assessment considers the relevance of the final pain points for each topic, and Painsight records the highest performance across the five product categories. Our method exhibits an average performance improvement of 0.35 points over LDA and 0.14 points over BERTopic. The results in Table \ref{tab:human evaluation} demonstrate the effectiveness of ITM, which could not be solely assessed using the NPMI metrics in Table \ref{tab:Iterative Topic Modification}. By employing Tasks 1, 2, and 3, the Painsight pipeline, which extracts pain points based on relevant tasks, also records higher performance compared to the baseline in human evaluation results. This validates the appropriateness of Painsight as an automatic framework for pain point detection.

\section{Conclusion}

In this study, we propose Painsight, a novel framework for automatically extracting and evaluating pain points from customer reviews. We address the under-explored problem of pain point detection and present a practical pipeline for real-world scenarios. By employing sentiment analysis and topic modeling, we identify sentiment-aware and topic-aware pain points that reflect customer perceptions and various types of discomfort. The final output is obtained by extracting the most important words or features from the data using a gradient-based attribution score. This score enables us to determine which words or features are most critical in influencing the model's decision-making process and utilize this information in the post-process to recognize more accurate and meaningful pain points. Experimental results demonstrate that Painsight outperforms existing models on five product group reviews, with human evaluation results indicating a high level of agreement compared to the baseline. Future work could involve incorporating diverse customer feedback and constructing a high-quality benchmark dataset to further validate and enhance the proposed approach.

\section*{Limitations}

In prior research \cite{salminen2022detecting}, several challenges have been identified in this field, such as noisy or low-quality data, semantic ambiguity, absence of standards, social desirability bias, and the requirement for human intervention. Our study aimed to tackle the challenge of detecting pain points and devised various strategies for managing noisy real-world reviews. Nonetheless, to fully unlock the potential of the Painsight, additional research is necessary to explore the wide range of emotional polarities beyond the generic `negative' sentiment. Furthermore, customer reviews often show mixed sentiments, which calls for addressing semantic ambiguity. Lastly, the performance of Painsight assessment was constrained to five product categories, highlighting the need for a comprehensive, high-quality benchmark encompassing diverse domains and performance evaluations across distinct categories.

\section*{Ethics Statement}

Throughout our human evaluation, we collected demographic details such as name, age, gender, and highest education level, after securing participants' consent and assuring them that their information would be exclusively utilized for research purposes. The results from the human evaluation were anonymized to protect participant confidentiality. The authors meticulously examined all customer reviews employed in the assessment, verifying the absence of any offensive or biased material. Participants took part in the evaluation for an estimated 40 minutes. They were compensated with a 5,000 KRW (equivalent to 3.7 USD) gift card, which was marginally above the Korean minimum wage during that period.


\bibliography{reference}
\bibliographystyle{acl_natbib}

\appendix

\section{Data Distribution}
\label{app:data distribution}

\begin{figure*}[ht]
\centering
\includegraphics[width=1\textwidth]{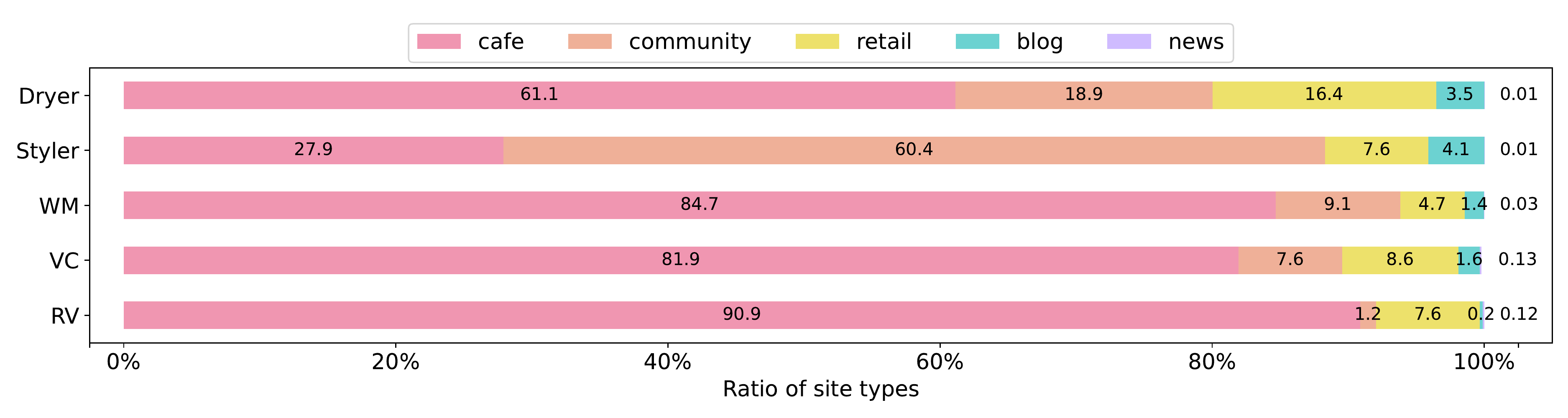}
\caption{Site type distribution of review data by product category.}
\label{fig:site type all}
\end{figure*}

The distribution of site types used for collecting review data for five product categories (Dryer, Styler, Washing Machine, Vacuum Cleaner, and Robotic Vacuum) is presented in Figure \ref{fig:site type all}. The data collection involved a diverse range of site types across product categories, with the highest number of reviews collected from cafe and community sites. The sentiment-aware pain point detection experiments utilized train, valid, and test datasets in an 8:1:1 ratio. However, the topic modeling in topic-aware pain point detection did not involve splitting the dataset.

\section{Training Details}
\label{sec:Training details}

The training process of Painsight was conducted using NVIDIA RTX 2080 Ti (for sentiment analysis and topic modeling) and RTX A6000 (for gradient-based attribution score and ITM). The PyTorch \footnote{\url{https://pytorch.org/}} library was performed for model training, and post-processing was carried out using Pororo \footnote{\url{https://github.com/kakaobrain/pororo}} and Captum \footnote{\url{https://captum.ai/}}. The hyperparameters used in Painsight are described in detail below:

\subsection{BERT + Vocab Expansion}
The model was trained using the transformers library \cite{wolf2019huggingface} with a BERT classifier. The batch size was set to 32, and the optimizer used was AdamW with a learning rate of 2e-05. The model was trained for 10 epochs, with 62 new vocabularies added. The maximum length of input sequences was set to 128.

\subsection{ITM}
The initial topic labels were generated using BERTopic, and the transformers library with a BERT classifier was used for classification. The optimizer used was Adam, and the model was trained for 1 epoch with a validation iteration of 10 and a maximum iteration of 100. Early stopping was used with patience 2. The maximum length of input sequences was set to 128, and the batch size was 64 with a learning rate of 3e-5.

\section{Evaluation Metric}

We employed Normalized Pointwise Mutual Information (NPMI) \cite{stevens-etal-2012-exploring} as an evaluation metric to measure the performance of ITM. NPMI is a widely used measure of the correlation between two words, which is computed by normalizing Pointwise Mutual Information (PMI). PMI measures the probability of two words occurring together, taking into account the frequency of their individual occurrences. However, PMI tends to overestimate the importance of infrequent words. To address this issue, NPMI normalizes PMI by considering the probability of the respective words. Through this normalization process, NPMI can more accurately measure the correlation between two words and ranges between -1 and 1.
NPMI is often used in topic modeling and is computed using the following formula:

\begin{align}
\text{PMI}(w_i, w_j) = {\frac{P(w_i, w_j)}{ P(w_i)P(w_j)}}, \label{eq:PMI} \\
\text{NPMI}(w_i, w_j) = {\frac{\text{PMI}(w_i, w_j)}{-\log (P(w_i, w_j))}}. \label{eq:NPMI}
\end{align}

Here, $P(w_i, w_j)$ denotes the probability of words $w_i$ and $w_j$ co-occurring, while $P(w_i)$ and $P(w_j)$ represent their individual probabilities. The numerator normalizes the probability of the two words occurring together by dividing it by the product of their individual occurrence probabilities. The denominator uses the log value of their co-occurrence probability to obtain PMI. Using NPMI, we can extract sets of related words in topic classes classified through ITM and evaluate if each topic has coherence. Therefore, we evaluated the consistency of each topic in LDA, BERTopic, and ITM with a set of related reviews using NPMI.

\section{Human Evaluation}

The evaluation instructions provided to annotators for each task in human evaluation are as follows, and the example is shown for the dryer.

\subsection{Task 1: Sentiment-aware pain point evaluation (Pain point candidates)}

Each sheet contains three items for each product group:
\begin{itemize}
\setlength\itemsep{-0.2em}
\item Consumer reviews of appliances for each product group collected online
\item Keywords extracted for each review (2-4 per review)
\end{itemize}

Please rate how well the keywords were extracted from each customer review:
\begin{itemize}
\setlength\itemsep{-0.2em}
\item 1: All keywords were extracted incorrectly.
\item 2: Keywords were generally not extracted.
\item 3: Keywords were extracted at an average level (50\% of all keywords).
\item 4: Keywords were generally well extracted.
\item 5: All keywords were extracted well.
\end{itemize}

\subsection{Task 2: Sentiment-aware pain point evaluation (Final pain points)}

\begin{figure}[t!]
\centering
\includegraphics[width=1\columnwidth]{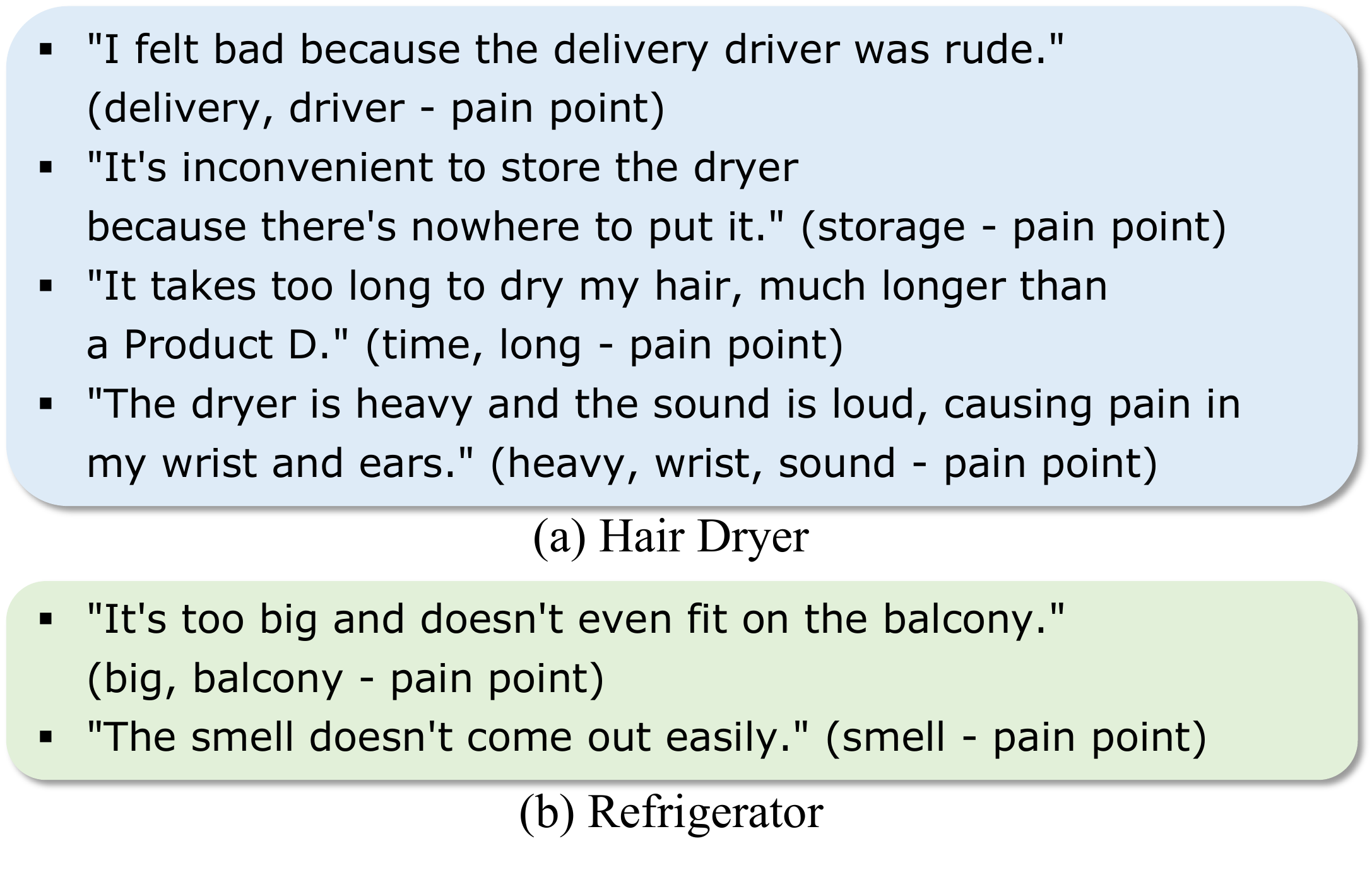}
\caption{Examples of Pain Points: We explained the concept of pain points using examples of home appliances that were not used in the evaluation, specifically hair dryers and refrigerators.}
\label{fig:example of painpoint}
\end{figure}

Evaluate the suitability of pain points identified as frequently appearing in negative reviews of the product category after reviewing 100 example sentences. To clarify the concept of "pain point," we provide example sentences (Figure \ref{fig:example of painpoint}):

Scoring criteria:
\begin{itemize}
\setlength\itemsep{-0.2em}
\item 0: This pain point cannot be considered a pain point for the product category.
\item 1: This pain point can be considered a pain point for the product category.
\end{itemize}

Notes:
\begin{itemize}
\setlength\itemsep{-0.2em}
\item Pain points may be composed of morphemes or spaced units rather than conventional word structures, selected based on their meaning.
\item There may be cases where a pain point cannot be extracted according to the logic. In such cases, you can give a score of 1.
\end{itemize}

\subsection{Task 3: Topic-aware pain point evaluation}

The following example contains three elements:
\begin{itemize}
\setlength\itemsep{-0.2em}
\item Consumer reviews on dryers collected from online community 
\item Review examples for each topic class
\item Key pain points for each topic class
\end{itemize}

20 reviews are provided for each topic class. Please assess the reviews in detail and evaluate whether the pain point can be used as a representative pain point for the topic class, using the following scores:
\begin{itemize}
\setlength\itemsep{-0.2em}
\item 0: The pain point cannot represent the topic class.
\item 1: The pain point can represent the topic class.
\end{itemize}

Notes:
\begin{itemize}
\setlength\itemsep{-0.2em}
\item Pain points can be constructed at the morpheme level, rather than a common word structure, depending on their meaning.
\item Do not consider overlaps in meaning or form between pain points. Please only judge the representativeness of each pain point for the reviews on the topic.
\item If the topic modeling performance is low, the topics of each review may not match. If the meaning of the topic cannot be identified through the sampled sentences for each topic class, you can assign 0 points to all the pain points.
\item Each topic often includes more than 500-1,000 sentences. Some pain points may be not contained in the sampled review. If similar words to that pain points, however, they could be suitable pain points for its cluster.  ([ex] "I tried it myself and my wrist hurts", pain point (`arms': considered correct))
\end{itemize}

\end{document}